# Large Language Models for Full-Text Methods Assessment: A Case Study on Mediation Analysis

Wenqing Zhang[1], BS; Trang Nguyen[1,2], PhD; Elizabeth A. Stuart[1,2], PhD; Yiqun T. Chen[1,3]*, PhD

[1] Department of Biostatistics, Johns Hopkins Bloomberg School of Public Health, Baltimore, MD 21205, United States
[2] Department of Mental Health, Johns Hopkins Bloomberg School of Public Health, Baltimore, MD 21205, United States
[3] Department of Computer Science, Johns Hopkins Whiting School of Engineering, Baltimore, MD 21218, United States

*Corresponding author: **Yiqun T. Chen, PhD**, Department of Biostatistics, Johns Hopkins Bloomberg School of Public Health, 615 N Wolfe St, Baltimore, MD 21205, United States.
Email: yiqunc@jhu.edu

**ABSTRACT:**

Systematic reviews are crucial for synthesizing scientific evidence but remain labor-intensive, especially when extracting detailed methodological information. Large language models (LLMs) offer potential for automating methodological assessments, promising to transform evidence synthesis. Here, using causal mediation analysis as a representative methodological domain, we benchmarked state-of-the-art LLMs against expert human reviewers across 180 full-text scientific articles. Model performance closely correlated with human judgments (accuracy correlation 0.71; F1 correlation 0.97), achieving near-human accuracy on straightforward, explicitly stated methodological criteria. However, accuracy sharply declined on complex, inference-intensive assessments, lagging expert reviewers by up to 15%. Errors commonly resulted from superficial linguistic cues—for instance, models frequently misinterpreted keywords like "longitudinal" or "sensitivity" as automatic evidence of rigorous methodological approaches—leading to systematic misclassifications. Longer documents yielded lower model accuracy, whereas publication year showed no significant effect. Our findings highlight an important pattern for practitioners using LLMs for methods review and synthesis from full texts: current LLMs excel at identifying explicit methodological features but require human oversight for nuanced interpretations. Integrating automated information extraction with targeted expert review thus provides a promising approach to enhance efficiency and methodological rigor in evidence synthesis across diverse scientific fields.

**Introduction**

Systematic reviews are widely considered the gold standard in evidence-based medicine, but conducting them takes considerable time and specialized knowledge. The rigorous process of comprehensively screening thousands of citations, evaluating each study's quality, and synthesizing findings can delay the translation of critical evidence into practice for years [1–3]. This challenge is especially pronounced in methodology reviews, such as those examining the use of statistical methods in the social and medical sciences. These reviews require careful examination of the text to evaluate which analytical models were applied in each study and the extent to which their underlying assumptions were stated and assessed [4,5].

Mediation analysis exemplifies this challenge. Widely used to investigate how an exposure affects an outcome through an intermediate variable (the mediator), mediation analysis rests on several key assumptions to make causal statements [6]. These include that (i) both the exposure and the mediator should influence the outcome; (ii) that the relationships follow the proper temporal sequence (exposure before mediator, mediator before outcome), (iii) that no unmeasured confounders bias the causal relationships among exposure, mediator, and outcome; (iv) sometimes parametric model assumptions [7]. Importantly, many of these assumptions are inherently untestable given the underlying counterfactual nature of the causal questions, and often require expert judgement to assess. Violations of these assumptions can lead to biased or misleading conclusions about the role of the mediator [7]. In practice, however, such assumptions are often overlooked or only partially addressed. When noted, they may be dispersed across sections, such as mentioned in the discussion or embedded in model diagnostics, making it difficult to judge whether the method has been applied appropriately.

Given the extensive expert time required for such reviews and recent advances in artificial intelligence (AI), there is growing interest in whether AI tools, particularly large language models (LLMs), can assist with or even automate the assessment of methodological assumptions in systematic reviews. LLMs have shown promise in related tasks such as identifying relevant papers and extracting statistics or conclusions from abstracts; however, most existing work to date has focused primarily on screening abstracts rather than analyzing full texts [8–15]. While some studies have explored full-text screening and AI-powered systematic reviews, these efforts mostly targeted straightforward features such as the number of participants and their recorded demographics, rather than more nuanced methodological assessments that demand domain expertise [16–18].

To address these gaps, we designed and evaluated an LLM-based pipeline to automate methodological quality assessment from full-text articles. We used mediation analysis as a case study and our evaluation targeted a set of methodological features with varying levels of complexity, and aims to approximate the role of a domain expert reviewer. We built on prior work showing that most high-impact psychology and psychiatry studies published between 2013 and 2018 fall short of methodological standards and rarely discuss the underlying assumptions of mediation analysis [7]. As part of that months-long manual review, the authors of that work collected an expert-annotated dataset from a completed manual review, which we used as ground truth. We evaluated whether LLMs can reliably perform expert-level methodological review by comparing their performance against human experts and summarizing detailed error patterns. Our evaluation framework provides a broadly applicable template for statistical methods assessments across diverse social and medical science domains.

## Methods

### Dataset Construction

We analyzed the same set of 180 articles analyzed by Stuart et al. [7], which were selected from the articles published between 2013 and 2018 in the 15 highest-impact psychiatry and psychology journals (ranked by impact factor at the time). These articles were identified through the PsycInfo database using the search term "mediation analys*".

In Stuart et al. [7], each article was independently evaluated by two expert reviewers (assigned from a panel of seven) on a set of methodological categories related to key assumptions in mediation analysis. Whenever reviewers disagreed on the answer, they engaged in a structured reconciliation process: each reviewer provided written, evidence-based justifications, followed by detailed discussions to determine whether disagreement stemmed from oversight, interpretive differences, or unclear article content. We considered 14 binary categories across the 180 articles and used the documented final consensus as our gold standard labels. Importantly, this consensus process also enables us to assess the accuracy of individual human reviewers against the consensus standard, under the assumption that at least one reviewer would have identified any methodological error or oversight.

To prepare the text data for computational analysis, we first retrieved full-text PDFs through automated downloads from publisher sites or PubMed Central. We then applied a state-of-the-art PDF extraction pipeline to convert each PDF into a text file, capturing all article content, including body text (sections, headings, and paragraphs) as well as tables [19]. Reference lists were excluded after extraction due to their limited relevance and to avoid exceeding the context window of the LLMs.

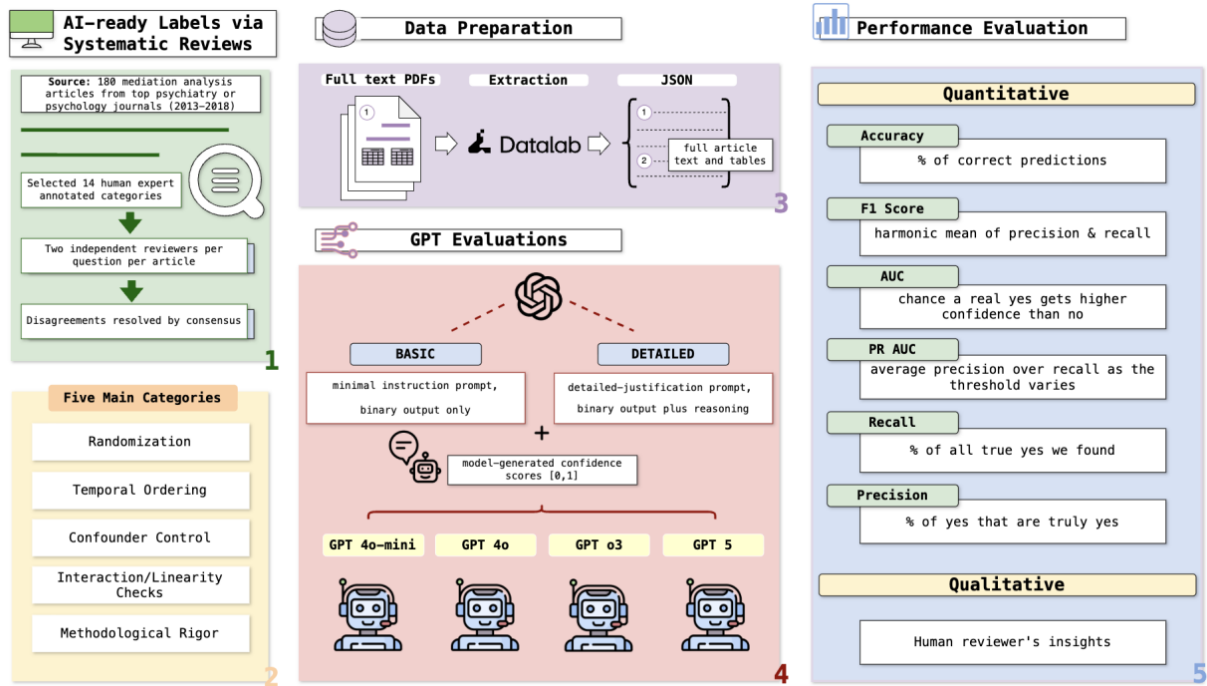

*Figure 1. Study workflow for evaluating large language model performance on causal mediation methodological assessment.* Article selection and expert labels were drawn from Stuart et al. [7], who reviewed 180 mediation analysis articles from top psychiatry and psychology journals and generated consensus labels on 14 methodological categories. We adopted these final consensus decisions as gold-standard reference labels. Building on this foundation, our study applied text parsing, large language model inference (basic vs. detailed prompts), and both quantitative and qualitative performance evaluation.

**Evaluation Framework**

| Abbrevated form | Full Question | Domain Each Belongs to |
|---|---|---|
| Random exposure | Was the exposure randomized? | Randomization |
| Causal Mediation | Did the study use causal mediation methods? | Methodological Rigor |
| Linearity Tests | If the Baron & Kenny approach was used, did the study assess whether linear relationships were appropriate for the mediator-outcome path? | Interaction and Linearity Assessments |
| Interaction Effects | If the Baron&Kenny approach was used, did the study examine for interaction between treatment and mediator on outcome? | Interaction and Linearity Assessments |

| Cov Exp-Med | Were covariates included in the exposure - mediator model? | Confounder Control |
|---|---|---|
| Cov Exp-Out | Were covariates included in the exposure - outcome model? | Confounder Control |
| Cov Med-Out | Were covariates included in the mediator - outcome model? | Confounder Control |
| Baseline Mediator | Whether the model adjusted for included baseline measures of the mediator? | Confounder Control |
| Baseline Outcome | Whether the model adjusted for included baseline measures of the outcome? | Confounder Control |
| Temporal Exp->Med | Was there temporal ordering of exposure before the mediator? | Temporal Ordering |
| Temporal Med->Out | Was there temporal ordering of mediators before the outcome? | Temporal Ordering |
| Assumption Discussion | Were mediation assumptions discussed? | Methodological Rigor |
| Sensitivity Analysis | Was sensitivity analysis to assumptions performed? | Methodological Rigor |
| Post-Exp Control | Does model control for other post-exposure variables? | Confounder Control |

*Table 1: Mediation Analysis Quality Assessment Criteria. Evaluation criteria adapted from Stuart et al. [7] for assessing adherence to key assumptions and best practices in mediation analysis. The framework encompasses 14 binary assessments across five domains: randomization, temporal ordering, confounder control, interaction and linearity assessments, and methodological rigor. Each criterion evaluates whether studies explicitly addressed fundamental requirements for valid causal inference in mediation analysis.*

We assessed each article against 14 binary methodological criteria aligned with Stuart et al.'s study [7], spanning five critical domains: randomization (randomized exposure assignment), temporal ordering (exposure precedes mediator; mediator precedes outcome), confounder control (covariates included in exposure-mediator, exposure-outcome, and mediator-outcome models; control for baseline mediator and outcome; control for post-exposure confounding variables), interaction and linearity assessments (examination of mediator-outcome linearity; evaluation of exposure-mediator interactions), and methodological rigor (discussion of mediation assumptions; sensitivity analyses for assumption violations; implementation of causal mediation methods); see **Table 1.** Each criterion required a binary assessment (Yes/No), yielding 2,520 total binary classifications per model run.

**Model Implementation**

In this study, we evaluated four state-of-the-art language models' (as of September 2025) ability to predict the binary labels across the 14 categories: GPT-4o [20], GPT-4o-mini [20], GPT-o3 [21], and GPT-5 [22]. To assess how prompt design influenced model performance, we implemented two complementary prompting strategies: BASIC prompts, which requested direct binary (Yes/No) responses without explanations, and DETAILED prompts, which included explanations and illustrative examples clarifying the meaning of each criterion (see **Appendix A.4** for the exact prompts used). Both strategies also instructed models to provide confidence scores [0,1] for each assessment, which we used to compute the area under the receiver operating characteristic curve (AUC) and precision-recall AUC (PR-AUC) in addition to accuracy metrics (see **Appendix A.1** for details).

We set temperature to T=1.0 for o3 and GPT-5 by model default, and T=0.5 for GPT-4o and GPT-4o-mini, based on preliminary testing to optimize the trade-off between response consistency and output diversity for each model. Results for GPT-4o, GPT-4o-mini, and GPT-o3 are all averaged over 5 runs; GPT-5 used single runs due to preview access limitations. For advanced reasoning models (i.e., GPT-5 and o3), the "chain-of-thought" reasoning process of asking the model to think before giving a prediction is enabled by default [23]. In a sensitivity analysis we conducted the same experiments with two additional non-ChatGPT models (Claude Sonnet 4 and Gemini 2.5) and found similar results; results are included in **Appendix A.3**.

**Statistical Analysis and Error Inspection**

We evaluated six performance metrics: accuracy, the fraction of correct predictions; precision, the proportion of predicted positives that are truly positive; recall, the proportion of true positives correctly identified; F1 score, the harmonic mean of precision and recall; AUC, the probability that a random positive is ranked above a random negative; and PR-AUC, which emphasizes performance on rare positive events. All metrics range from 0 to 1, with higher values indicating better performance.

To establish a benchmark of human-level performance, we assessed each individual reviewer's performance against the gold standard labels. Specifically, we computed accuracy, precision, recall, and F1 score for each individual reviewer based on each reviewer's initial pre-reconciliation assessments. This provides context for interpreting LLM performance relative to human experts. Note that we did not compute AUC and PR-AUC for the human benchmark, as these metrics require continuous confidence scores, which were not provided by reviewers in the initial study.

To better understand factors associated with LLM accuracy, we also regressed article-level performance on article characteristics including full-text word count, structural features (e.g., presence of clearly labeled Methods section), and publication year. Finally, to probe model behavior, we examined error patterns through human inspection analyses, identifying whether failures arose from knowledge gaps, study-design misinterpretation, or inconsistent reasoning. All code to reproduce analyses is available at https://github.com/yiqunchen/CausalJudge.

**Results**

Across 14 methodological criteria in 180 full-text articles, individual human reviewers (before reconciliation) averaged 88.8% accuracy and 57.1% F1 when compared to the gold standard labels. Advanced reasoning LLMs (GPT-5 and GPT-o3) achieved comparable performances, which achieved 87.5% accuracy and 64.0% F1. Human reviewers showed slightly higher overall accuracy, whereas LLMs achieved higher F1 scores, partly because human recall was lower for criteria with sparse positives—such as linearity, sensitivity analysis, and post-exposure control—which consequently reduced human F1 scores. Overall, there is a high correlation between human and LLM metric scores (**Figure 2**), suggesting that harder questions for humans are likely posing challenges for LLMs too, mirroring recent research findings in LLM reasoning abilities [24]. Moreover, human performance also varied substantially across reviewers: the top-performing reviewer achieved near-perfect performance on almost all criteria, with an average accuracy of 97.7% and F1 of 89.3%, except for linearity tests, where the F1 fell to 0.17. We present LLM performance alongside the aspirational benchmark set by the best-performing human reviewer in the main text, with results for the remaining six reviewers available in **Appendix A.2**.

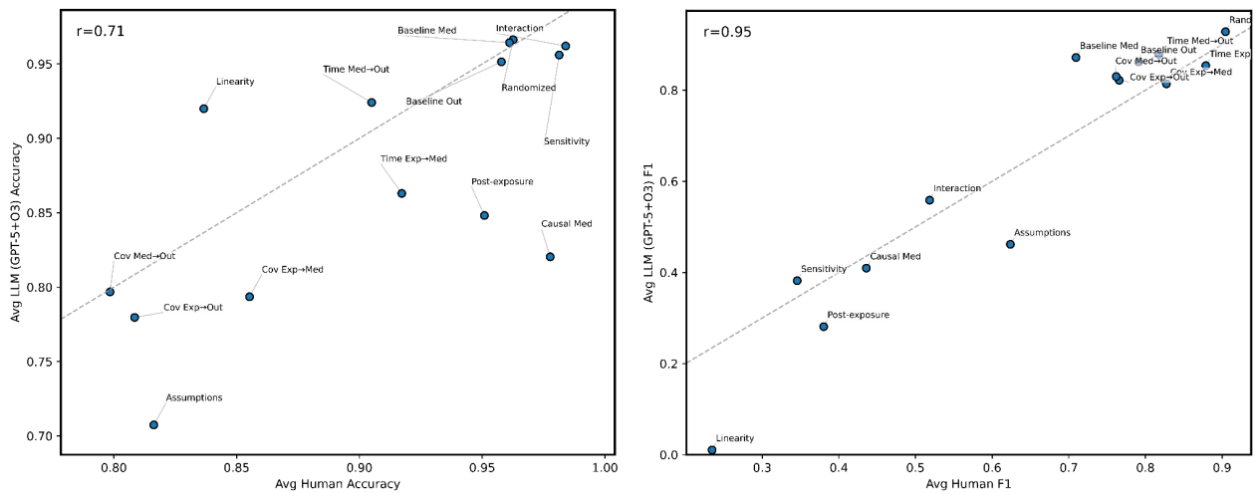

*Figure 2. Average human reviewer performance versus advanced reasoning LLMs (GPT-5 and GPT-o3), measured by accuracy (left) and F1 (right). Both panels show a strong Pearson correlation between human and LLM performance.*

**Performance by Methodological Domain**

Because the prevalence of studies meeting each methodological criterion varied widely across the 14 categories (see **Table 1**), the resulting binary classification problems exhibited varying degrees of class imbalance. To account for this imbalance, we evaluated performance using both accuracy (overall correctness) and F1 score (the model's effectiveness at identifying rare positives). This dual evaluation strategy enabled us to highlight performance not only in terms of overall correctness but also sensitivity to less frequent methodological features. Using these metrics, we observed substantial performance variation across the five methodological domains, particularly evident in the F1 scores.

**Randomization** (1 criterion: randomized exposure assignment). All models achieved near-human accuracy (0.95–0.97) on identifying whether the exposure was randomized, with minimal benefit from detailed prompting. Even GPT-4o-mini attained an F1 of 0.91–0.92, approaching the human benchmark (1.00).

**Temporal Ordering** (2 criteria: exposure precedes mediator; mediator precedes outcome). Reasoning-based models (GPT-5 and GPT-o3) performed strongly here, achieving high accuracy (0.86–0.93) and consistent F1 scores (0.85–0.88). GPT-4o showed mixed performance (accuracy 0.61–0.75; F1 0.61–0.76), while GPT-4o-mini struggled significantly, particularly on whether the mediator precedes the outcome temporally (accuracy: 0.34). Prompt engineering provided only modest improvements (F1 gains: ~0.01–0.03; accuracy gains: ~0.01–0.06).

**Confounder Control** (6 criteria across covariates in three model relationships, baseline mediator/outcome control, and post-exposure control). Determining whether the model adjusted for baseline measures of the mediator or outcome generally showed high but model-dependent performance: GPT-5 and GPT-o3 closely approaching human accuracy (0.94–0.97), while GPT-4o-mini consistently performed more poorly. Identifying whether the model included post-exposure controls proved particularly challenging: despite high accuracy (>0.86), F1 scores were low (<0.37) due to poor precision (LLMs: 0.15–0.30 vs. the best human reviewer: 0.89).

**Interaction and Linearity Assessments** (2 criteria: mediator-outcome linearity and exposure-mediator interactions). Evaluations of linearity assumptions (whether mediator-outcome relationships were appropriately linear) and interaction effects (whether studies examined exposure-mediator interactions) presented significant challenges. Severe class imbalance, characterized by the rarity of explicit linearity and interaction assessments in original studies, led to extremely low F1 scores (<0.10 for LLMs, 0.17 for humans) despite high accuracy (0.91–0.92). Reasoning models (GPT-5, GPT-o3) managed slightly better performance on interaction effects, reaching moderate F1 scores (0.50–0.60 compared to 0.89 for the best human reviewer), whereas non-reasoning models showed significantly weaker outcomes. These accuracy-F1 discrepancies underscored the limitations of accuracy alone in scenarios marked by class imbalance.

**Methodological Rigor** (3 criteria: assumption discussion; sensitivity analyses; causal mediation implementation). When it comes to whether studies explicitly discussed mediation assumptions, conducted sensitivity analyses, or applied formal causal mediation methods—prompting strategies had the most pronounced effects empirically. Specifically, detailed prompting significantly improved performance in identifying studies that explicitly used causal mediation methods (GPT-5: F1 +0.30; GPT-o3: F1 +0.35), alongside notable accuracy improvements. However, detailed prompts had minimal impact on whether studies discussed underlying assumptions or performed sensitivity analyses, with only GPT-o3 showing modest improvement (+0.10 F1).

**Summary of Cross-Domain Patterns.**

Three trends emerged across the 14 criteria. First, LLM performance was strongest for explicitly stated criteria (randomization, covariates), moderate for structural requirements (temporal ordering, baseline adjustments), and weakest for inference-heavy tasks (linearity, post-exposure controls, sensitivity analyses). Second, detailed prompting yielded targeted rather than uniform gains, most notably improving causal mediation identification. Third, substantial divergence between accuracy and F1 scores on complex, imbalanced criteria (e.g., linearity, post-exposure control, sensitivity analyses) highlighted the need for complementary metrics.

**Paper Characteristics and Model Error Analysis**

Analysis of paper characteristics indicated that LLMs struggled to assess methodological quality in longer articles. Articles in the top accuracy quintile (highest 20%) averaged only 623 words, whereas those in the bottom accuracy quintile (lowest 20%) averaged significantly longer at 3,089 words (two-sample t-test; p<0.001). This aligns with established limitations of current LLMs, where longer texts strain context windows and attention mechanisms, thereby reducing performance [25,26]. Conversely, LLM performance did not differ significantly between papers with and without an explicit Methods section (p=0.8), suggesting that LLMs effectively identified methodological details even without explicit structural cues.

We also found interesting error patterns. Although reasoning models like GPT-5 achieved the high overall accuracy comparable to average human reviewers, most of its mistakes involve overinterpreting weak or misleading linguistic cues. One notable example is the presence of the word "*experiment*". For instance, the model inferred a false positive for randomized exposure from the phrase *"In two experiments, we investigated the effect of social distance"*, despite the study actually employing a within-subjects design manipulated via instruction rather than random assignment [27]. Another example is in a juvenile-justice study, where the authors drew data from an ongoing randomized clinical trial but only analyzed a cross-sectional sample at baseline so there is no randomization; GPT-5 again inferred a randomized design for the analysis [28].

Older models such as GPT-4o showed more technical misunderstandings. For instance, it often misclassified routine demographic adjustments—such as "adjusted for age, sex, and depressive mood"—as control for post-exposure variables, when these covariates were actually pre-exposure factors [29]. Across all models, accurately identifying temporal ordering and sensitivity analysis remained universal challenges, as these concepts are described inconsistently and appear in varied contexts across articles.

Identifying correct temporal ordering among exposure, mediator, and outcome proved particularly challenging across the evaluated LLMs. As seen in other domains, the mere presence of words like "temporal" frequently resulted in false-positive assessments. For example, language models incorrectly interpret statements such as "both emotion regulation and dissociation were assessed at the same timepoint (T1), and thus the temporal sequencing of mediators remains to be ascertained through a longitudinal design" [30] as explicit confirmation of temporal sequencing of exposure → mediator → outcome. This reflects a broader pattern where LLMs equate the presence of longitudinal language in literature reviews or limitation sections with an actual longitudinal design implemented in the reported analyses.

Finally, the mentioning of "sensitivity analysis" was frequently over-credited as rigorous causal stress-testing. Studies performing simple sensitivity checks—for example, "As the cut-offs for caseness could influence subgroup effects, we performed sensitivity analyses using alternative cut-offs (0–11, 12–19, and 20+)"—were incorrectly credited by models as meeting causal identification standards, thus overstating inferential rigor [31]. Collectively, these patterns reflect a reliance on superficial linguistic cues in the paper (e.g., "randomized", "temporal", "sensitivity") rather than reasoning based on the actual study design and statistical analyses described in the papers

**Discussion**

Our evaluation of GPT-family models in extracting methodological details from mediation analysis papers reveals a clear performance gradient. Models excel at identifying explicitly stated methodological features, such as randomization and covariate inclusion. Performance is intermediate for structured requirements like temporal ordering: reasoning models (GPT-5 and GPT-o3) show generally strong performance but remain sensitive to variations in phrasing, whereas GPT-4o and GPT-4o-mini consistently lag behind. By contrast, models perform more poorly on reasoning-heavy criteria (e.g., post-exposure control, sensitivity analyses, exposure–mediator interactions) and criteria with sparse positives (linearity tests). For these more complex tasks, even advanced models struggled to achieve high F1 scores and accuracy, especially compared to the best human reviewer. Nonetheless, overall model performance aligned closely with average human reviewers, as demonstrated by the strong correlation (F1 $r = 0.97$; details in appendix) between average model and human reviewer scores across the 14 criteria. This suggests that while language models may overly rely on superficial linguistic cues, human reviewers may share similar tendencies.

We also highlight the importance of careful evaluation. As in many LLM benchmarks, class imbalance is a serious concern in our dataset: depending on the distribution of positives and negatives, a naïve classifier that always predicts one class could achieve deceptively high accuracy. To mitigate this, we recommend complementary metrics less sensitive to class imbalance, such as F1 scores and PR-AUC, which provide better discriminatory power when benchmarking models against human reviewers.

Our study has several limitations. First, it specifically focuses on synthesizing methodological details of mediation analysis; thus, our findings might not translate directly to other methodological domains. Second, our evaluation relied on binary labels, whereas more nuanced, open-ended assessments could enhance the applicability of our findings to broader methodological synthesis. Additionally, comparative evaluations of this nature depend heavily on gold-standard reference labels, which are seldom accessible in practice. Expanding the evaluation framework to previously unassessed methodological reviews and incorporating exploratory (rather than solely confirmatory) assessments could further strengthen this work. Moreover, while GPT-family models and comparably performant non-GPT models (such as

Claude 4 and Gemini 2.5; see **Appendix A.4**) showed similar performance patterns, future studies could systematically investigate whether reasoning models inherently outperform others on such tasks as well as exploring fine-tuning of open-weight models. Despite the specific focus on mediation analysis, the methodological criteria evaluated here—such as sensitivity analyses and temporal ordering—are broadly relevant and widely applicable, appearing in frameworks such as difference-in-differences and path analysis, thus contributing substantially to causal inference research across social and medical sciences.

**Conclusions**

Our study demonstrates that as of September 2025, state-of-the-art LLMs can match average human reviewers in extracting straightforward methodological information from systematic reviews. However, they still lag behind on nuanced, reasoning-intensive tasks, such as determining whether studies assessed exposure–mediator interactions or adequately controlled for post-exposure confounding. Notably, even the most advanced LLM models tested remain outperformed by the best human reviewers, highlighting the importance of criterion-specific deployment strategies. Specifically, LLM-only extraction may suffice for simpler, explicit methodological features, while complex or subjective assessments should involve human-AI collaboration. Our evaluation framework and pipeline are readily generalizable, providing a robust foundation for reproducible benchmarking and facilitating responsible integration of LLMs into diverse evidence-synthesis workflows.

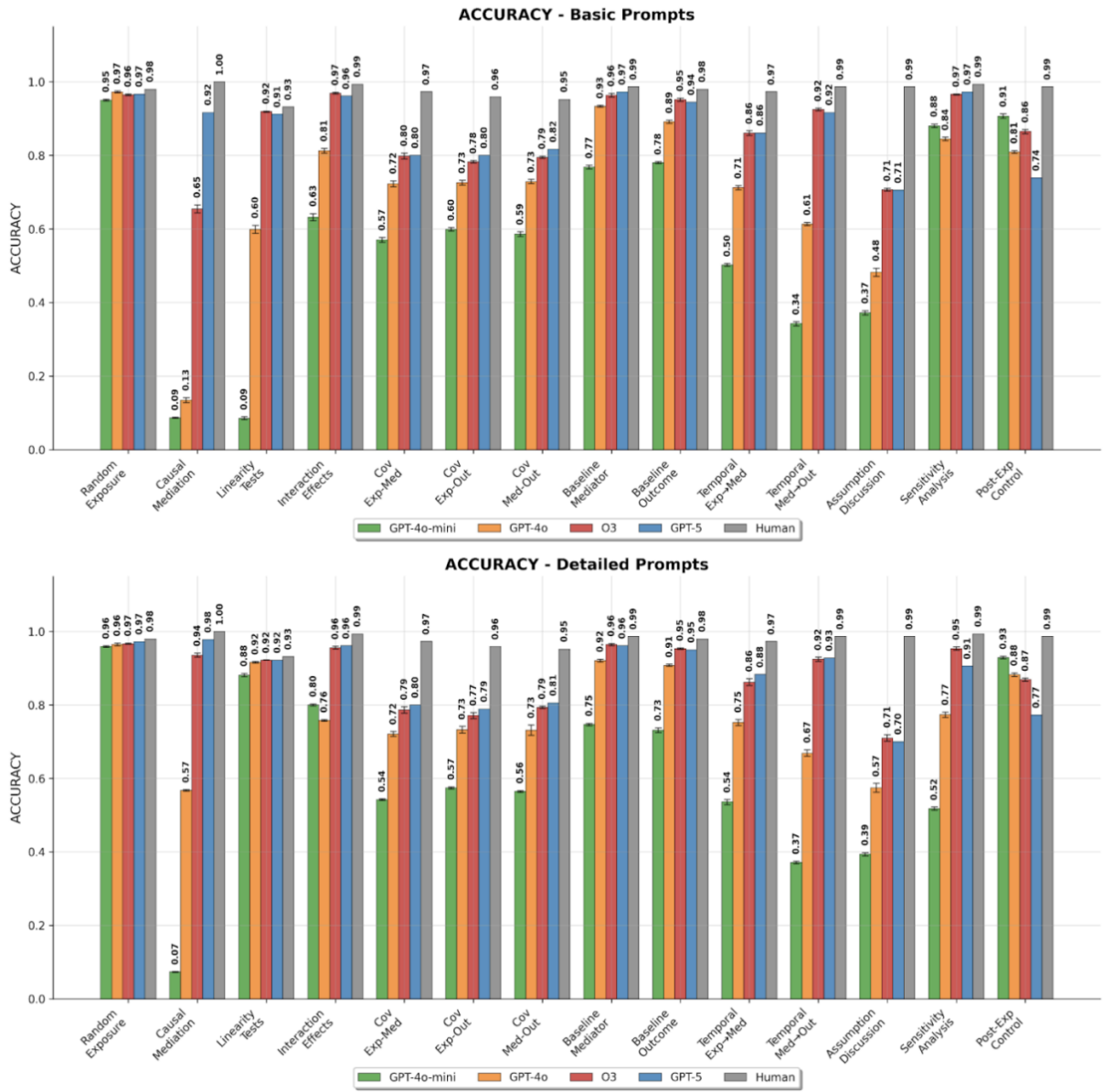

Figure 3. Model Accuracy Under Basic (Top) and Detailed (Bottom) Prompting Strategies Across 14 Methodological Criteria; colors represent different ChatGPT models and best human reviewer.

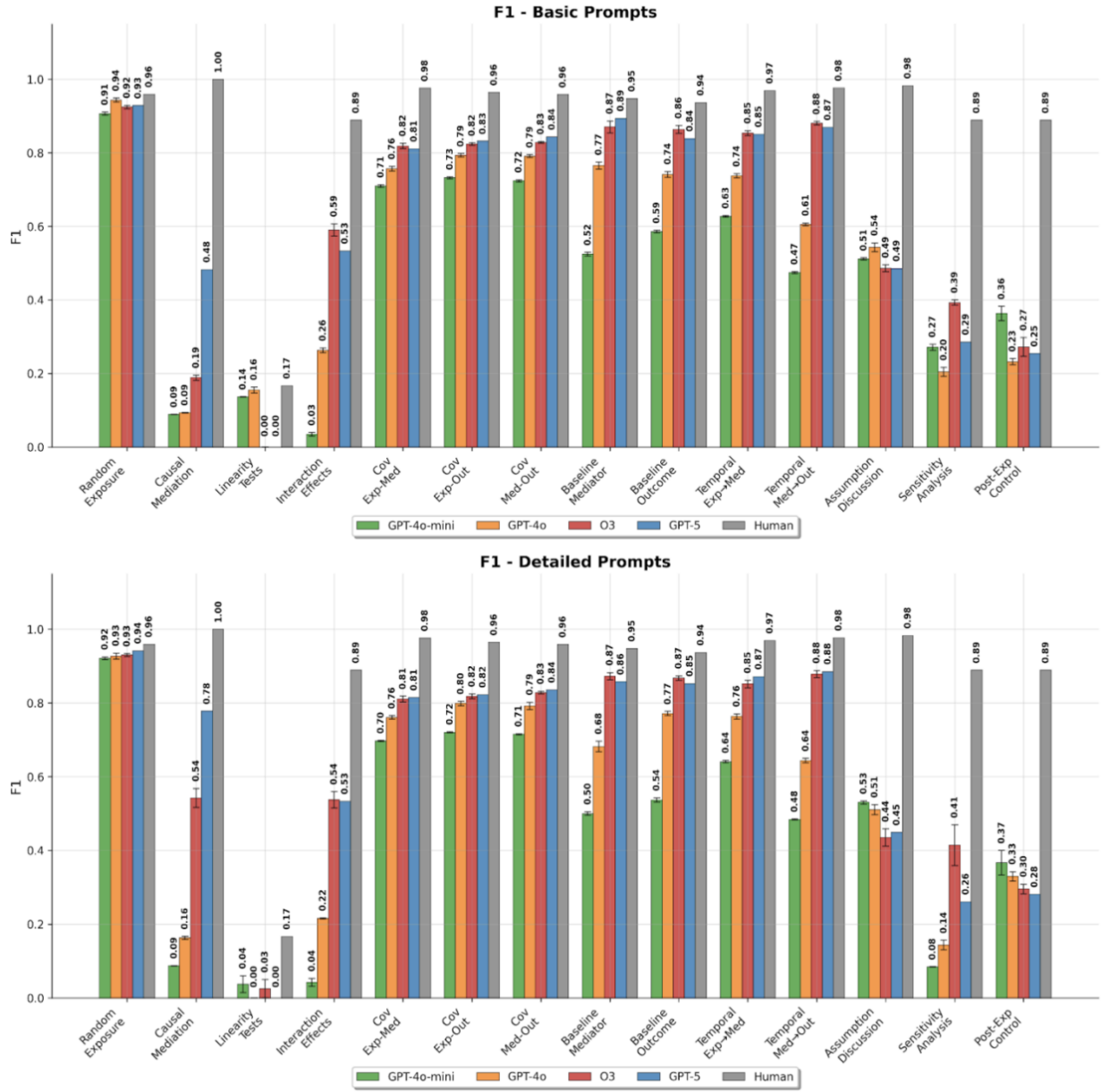

Figure 4. Model F1 Score Under Basic (Top) and Detailed (Bottom) Prompting Strategies Across 14 Methodological Criteria; colors represent different ChatGPT models and best human reviewer.

# Reference

1   Affengruber L, van der Maten MM, Spiero I, *et al.* An exploration of available methods and tools to improve the efficiency of systematic review production: a scoping review. *BMC Med Res Methodol*. 2024;24:210. doi: 10.1186/s12874-024-02320-4

2   Borah R, Brown AW, Capers PL, *et al.* Analysis of the time and workers needed to conduct systematic reviews of medical interventions using data from the PROSPERO registry. *BMJ Open*. 2017;7:e012545. doi: 10.1136/bmjopen-2016-012545

3   Tsertsvadze A, Chen Y-F, Moher D, *et al.* How to conduct systematic reviews more expeditiously? *Syst Rev*. 2015;4:160. doi: 10.1186/s13643-015-0147-7

4   Bouwmeester W, Zuithoff NPA, Mallett S, *et al.* Reporting and Methods in Clinical Prediction Research: A Systematic Review. *PLOS Med*. 2012;9:e1001221. doi: 10.1371/journal.pmed.1001221

5   Higgins J, Thomas J. *Cochrane Handbook for Systematic Reviews of Interventions*. Version 6.5 (updated August 2024). Cochrane 2024.

6   Imai K, Keele L, Tingley D. A general approach to causal mediation analysis. *Psychol Methods*. 2010;15:309–34. doi: 10.1037/a0020761

7   Stuart EA, Schmid I, Nguyen T, *et al.* Assumptions Not Often Assessed or Satisfied in Published Mediation Analyses in Psychology and Psychiatry. *Epidemiol Rev*. 2021;43:48–52. doi: 10.1093/epirev/mxab007

8   Delgado-Chaves FM, Jennings MJ, Atalaia A, *et al.* Transforming literature screening: The emerging role of large language models in systematic reviews. *Proc Natl Acad Sci*. 2025;122:e2411962122. doi: 10.1073/pnas.2411962122

9   Dennstädt F, Zink J, Putora PM, *et al.* Title and abstract screening for literature reviews using large language models: an exploratory study in the biomedical domain. *Syst Rev*. 2024;13:158. doi: 10.1186/s13643-024-02575-4

10  Issaiy M, Ghanaati H, Kolahi S, *et al.* Methodological insights into ChatGPT's screening performance in systematic reviews. *BMC Med Res Methodol*. 2024;24:78. doi: 10.1186/s12874-024-02203-8

11  Li M, Sun J, Tan X. Evaluating the effectiveness of large language models in abstract screening: a comparative analysis. *Syst Rev*. 2024;13:219. doi: 10.1186/s13643-024-02609-x

12  Oami T, Okada Y, Nakada T. Performance of a Large Language Model in Screening Citations. *JAMA Netw Open*. 2024;7:e2420496–e2420496. doi: 10.1001/jamanetworkopen.2024.20496

13  Pham B, Jovanovic J, Bagheri E, *et al.* Text mining to support abstract screening for knowledge syntheses: a semi-automated workflow. *Syst Rev*. 2021;10:156. doi: 10.1186/s13643-021-01700-x

14 Luo X, Tahabi FM, Marc T, *et al.* Zero-shot learning to extract assessment criteria and medical services from the preventive healthcare guidelines using large language models. *J Am Med Inform Assoc*. 2024;31:1743–53. doi: 10.1093/jamia/ocae145

15 Scherbakov D, Hubig N, Jansari V, *et al.* The emergence of large language models as tools in literature reviews: a large language model-assisted systematic review. *J Am Med Inform Assoc*. 2025;32:1071–86. doi: 10.1093/jamia/ocaf063

16 Khraisha Q, Put S, Kappenberg J, *et al.* Can large language models replace humans in systematic reviews? Evaluating GPT-4's efficacy in screening and extracting data from peer-reviewed and grey literature in multiple languages. *Res Synth Methods*. 2024;15:616–26. doi: 10.1002/jrsm.1715

17 Schopow N, Osterhoff G, Baur D. Applications of the Natural Language Processing Tool ChatGPT in Clinical Practice: Comparative Study and Augmented Systematic Review. *JMIR Med Inf*. 2023;11:e48933. doi: 10.2196/48933

18 Sercombe J, Bryant Z, Wilson J. Evaluating a Customized Version of ChatGPT for Systematic Review Data Extraction in Health Research: Development and Usability Study. *JMIR Form Res*. 2025;9:e68666. doi: 10.2196/68666

19 Platform. https://www.datalab.to/platform (accessed 11 September 2025)

20 OpenAI, Hurst A, Lerer A, *et al.* GPT-4o System Card. 2024.

21 Introducing OpenAI o3 and o4-mini. https://openai.com/index/introducing-o3-and-o4-mini/ (accessed 22 September 2025)

22 GPT-5 is here. https://openai.com/gpt-5/ (accessed 22 September 2025)

23 Wei J, Wang X, Schuurmans D, *et al.* Chain-of-thought prompting elicits reasoning in large language models. *Proceedings of the 36th International Conference on Neural Information Processing Systems*. Red Hook, NY, USA: Curran Associates Inc. 2022:24824–37.

24 Lampinen AK, Dasgupta I, Chan SCY, *et al.* Language models, like humans, show content effects on reasoning tasks. *PNAS Nexus*. 2024;3:pgae233. doi: 10.1093/pnasnexus/pgae233

25 Bai Y, Lv X, Zhang J, *et al.* LongBench: A Bilingual, Multitask Benchmark for Long Context Understanding. In: Ku L-W, Martins A, Srikumar V, eds. *Proceedings of the 62nd Annual Meeting of the Association for Computational Linguistics (Volume 1: Long Papers)*. Bangkok, Thailand: Association for Computational Linguistics 2024:3119–37.

26 Liu NF, Lin K, Hewitt J, *et al.* Lost in the Middle: How Language Models Use Long Contexts. *Trans Assoc Comput Linguist*. 2024;12:157–73. doi: 10.1162/tacl_a_00638

27 Zhang X, Liu Y, Chen X, *et al.* Decisions for Others Are Less Risk-Averse in the Gain Frame and Less Risk-Seeking in the Loss Frame Than Decisions for the Self. *Front Psychol*. 2017;8:1601. doi: 10.3389/fpsyg.2017.01601

28  Balazs J, Miklósi M, Keresztény Á, *et al.* Attention-deficit hyperactivity disorder and suicidality in a treatment naïve sample of children and adolescents. *J Affect Disord*. 2014;152–154:282–7. doi: 10.1016/j.jad.2013.09.026

29  Lee K, Lee H-K, Kim SH. Temperament and character profile of college students who have suicidal ideas or have attempted suicide. *J Affect Disord*. 2017;221:198–204. doi: 10.1016/j.jad.2017.06.025

30  Hébert M, Langevin R, Oussaïd E. Cumulative childhood trauma, emotion regulation, dissociation, and behavior problems in school-aged sexual abuse victims. *J Affect Disord*. 2018;225:306–12. doi: 10.1016/j.jad.2017.08.044

31  Li R, Cooper C, Barber J, *et al.* Coping strategies as mediators of the effect of the START (strategies for RelaTives) intervention on psychological morbidity for family carers of people with dementia in a randomised controlled trial. *J Affect Disord*. 2014;168:298–305. doi: 10.1016/j.jad.2014.07.008

**Appendix**

**Appendix A1: Additional Performance Metrics for LLMs**

We present AUC, PR-AUC, recall, and precision for the experiments run in the Methods section of the main text (Figures S1-S4).

**Performance by Methodological Domain**

**Randomization** (1 criterion: randomized exposure assignment).

All models achieved strong classification performance (recall 0.89–0.99, precision 0.84–0.98), with detailed prompting showing minimal impact on these metrics. GPT-4o exhibited higher discrimination under basic prompting (AUC 0.62) than other models (0.22–0.36); this difference narrowed with detailed prompting (GPT-4o 0.49; others 0.33–0.44). PR-AUC values were modest across models (0.19–0.40), consistent with <20% prevalence for this criterion and score-calibration differences; despite this, operating-point precision and recall remained high.

**Temporal Ordering** (2 criteria: exposure precedes mediator; mediator precedes outcome).

Both criteria showed high recall across models (0.91 and 1.00), approaching human performance (0.97 and 0.98). Precision varied substantially, driving performance stratification. For exposure→mediator ordering, GPT-5 achieved the highest precision (0.84, detailed prompting), while GPT-4o-mini lagged (0.48), both below the best human precision (0.97). Discrimination patterns differed by model and prompt: GPT-4o maintained AUC 0.85–0.88 across prompts and reached recall 1.00 for mediator→outcome with detailed prompting. GPT-o3 started strong (AUC 0.92 for mediator→outcome) but showed lower PR-AUC with detailed prompting (exposure→mediator 0.76→0.62; mediator→outcome 0.77→0.61), suggesting calibration and prompting sensitivity.

**Confounder Control** (6 criteria: covariates in three model relationships; baseline mediator/outcome control; post-exposure control).

This largest domain showed heterogeneous patterns. Covariate-inclusion checks were the most consistent across models. For exposure–mediator covariates, all models achieved high recall (0.81–1.00); GPT-o3 showed strong discrimination (AUC 0.80–0.82) and precision (0.76–0.78) across prompts. GPT-5 maintained competitive precision (0.80–0.81) with lower AUC (0.54–0.58). GPT-4o-mini reached very high recall with lower precision (0.54–0.55). Similar patterns held for exposure–outcome and mediator–outcome covariates, with GPT-o3 yielding the highest AUC (0.79–0.83) and precision (0.73–0.77), followed by GPT-5.

For baseline mediator control, GPT-4o had the highest discrimination (AUC 0.84–0.90) and strong precision under basic prompting (0.70), while detailed prompting reduced recall (0.85→0.66). GPT-o3 was more stable across prompts (AUC 0.74–0.85; precision 0.80–0.81). GPT-5 and GPT-4o-mini showed weaker discrimination despite high recall. Baseline outcome control followed similar patterns: GPT-4o achieved the highest AUC (0.96–0.97), whereas GPT-5 and GPT-o3 demonstrated higher precision (0.76–0.79).

Post-exposure control remained the most challenging criterion. Despite moderate AUC (0.48–0.73), precision was low across models (0.15–0.37) relative to the best human reviewer (0.89). GPT-5 paired slightly higher recall (up to 0.80) with low precision (0.15–0.17). PR-AUC was correspondingly low (0.06–0.29), reflecting class imbalance and a tendency to over-call positives on this criterion.

**Interaction and Linearity Assessments** (2 criteria: mediator-outcome linearity; exposure-mediator interactions).

Linearity showed clear limitations across models. Despite moderate AUC (0.46–0.57), practical classification performance was low. GPT-4o-mini's recall decreased with detailed prompting (0.93→0.03), and GPT-o3/GPT-5 had near-zero recall (0.00–0.01) under both prompts. Under basic prompting, only GPT-4o-mini (precision 0.07) and GPT-4o (0.09) identified any positives, below the best human precision (0.50). PR-AUC remained low (0.07–0.16), consistent with sparse positives; human recall was also low (0.10).

Interaction effects showed improved but still limited performance. GPT-4o achieved recall 1.00 across prompts with low precision (0.15 basic; 0.12 detailed), indicating over-identification. GPT-o3 provided the best balance (precision 0.53 basic; 0.42 detailed; AUC 0.92 basic; 0.88 detailed), though still below the best human precision (1.00). GPT-5 was moderate (precision 0.44), and GPT-4o-mini was low (precision 0.02–0.03; recall 0.13–0.20).

**Methodological Rigor** (3 criteria: assumption discussion; sensitivity analyses; causal mediation implementation).

Assumption discussion displayed heterogeneous patterns. GPT-4o-mini maintained high recall (0.94–0.98) across prompts; GPT-o3 and GPT-5 had lower recall (0.32–0.40) relative to the best human reviewer. GPT-o3 achieved the highest precision (0.63 basic; 0.68 detailed), above GPT-4o-mini (0.35–0.36) but below the best human (0.97). Discrimination was moderate (AUC 0.48–0.70), with GPT-o3 showing the highest PR-AUC (0.52–0.56).

Sensitivity analyses were challenging. GPT-4o-mini achieved recall 1.00 under both prompts; GPT-5 improved with detailed prompting (0.25→0.75). Precision, however, remained low across models (0.04–0.33) relative to the best human (0.80). GPT-4o showed the strongest discrimination (AUC 0.90–0.93) but low PR-AUC (0.13–0.20), consistent with class imbalance.

Causal mediation implementation had high recall across models (0.85–1.00) with lower precision (0.05–0.70) than the best human reviewer. GPT-5 improved with detailed prompting (precision 0.33–0.70); other models showed smaller gains. GPT-o3 yielded the highest discrimination (AUC 0.86–0.89; PR-AUC 0.54–0.56) with moderate precision (0.11 basic; 0.40 detailed).

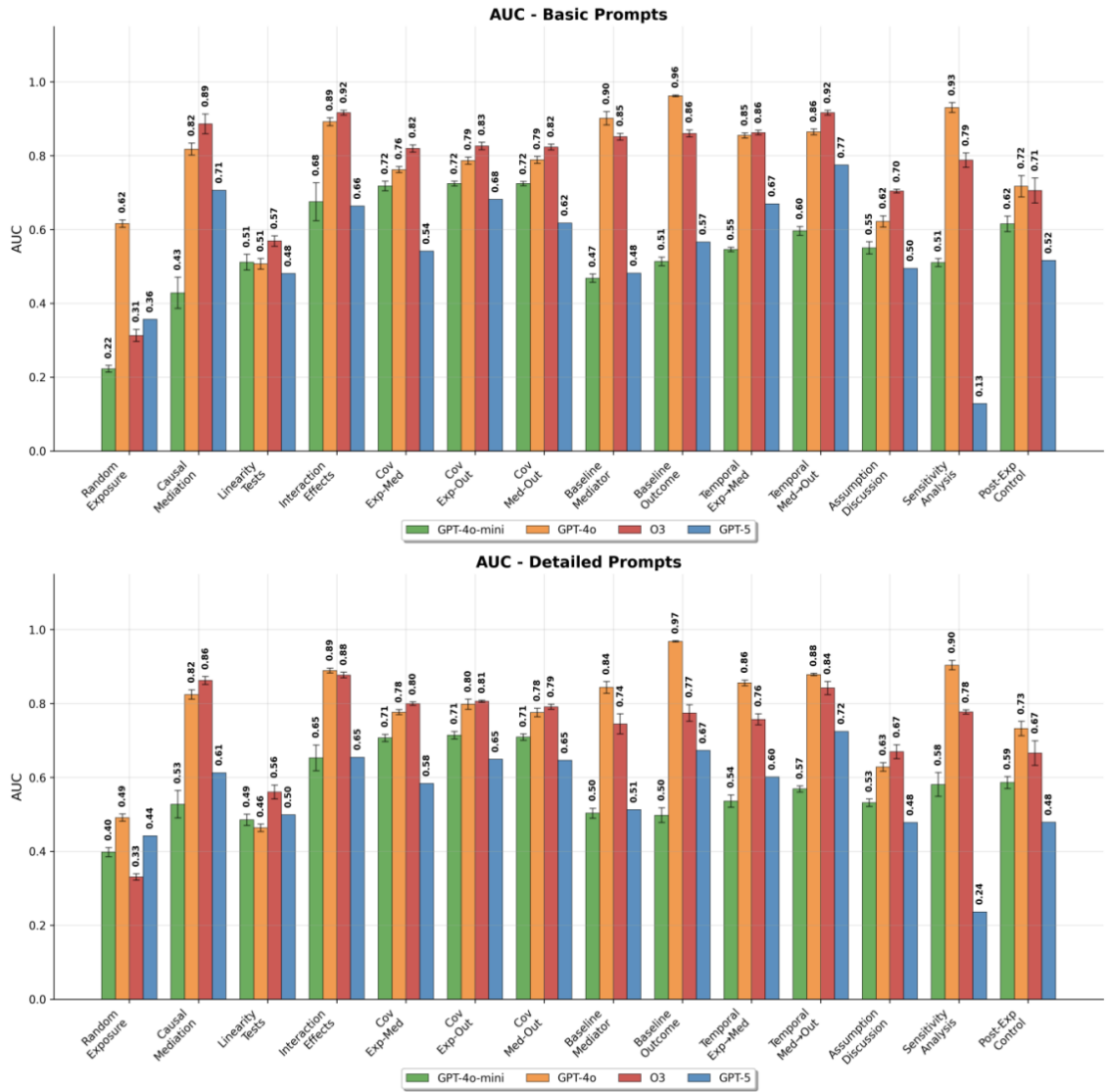

Figure S1. Model AUC Score Under Basic (Top) and Detailed (Bottom) Prompting Strategies Across 14 Methodological Criteria; colors represent different ChatGPT models. No human review as AUC requires continuous confidence score, which human reviewers don't have.

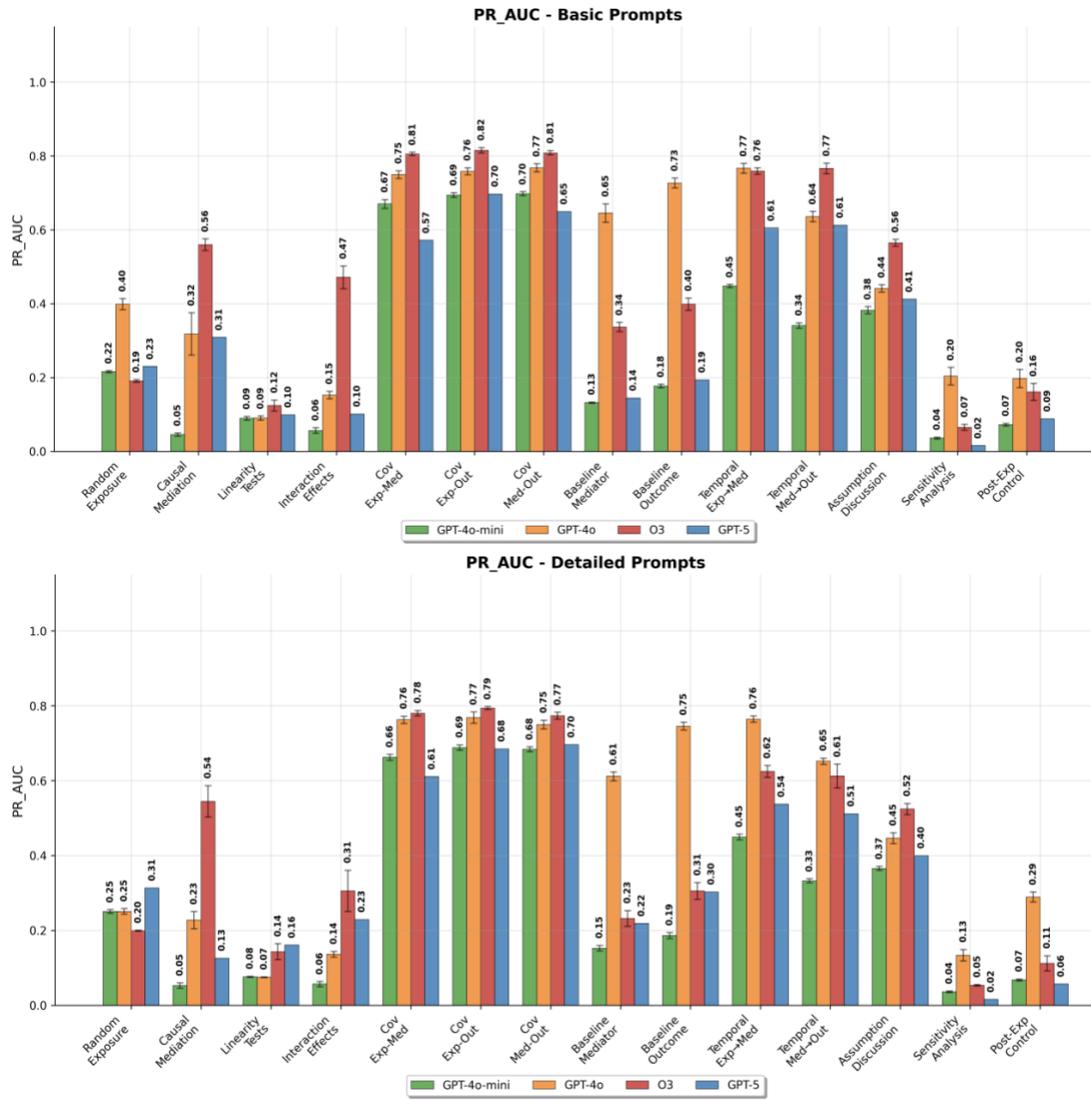

Figure S2. Model PR-AUC Score Under Basic (Top) and Detailed (Bottom) Prompting Strategies Across 14 Methodological Criteria; colors represent different ChatGPT models. No human review as AUC requires continuous confidence score, which human reviewers don't have.

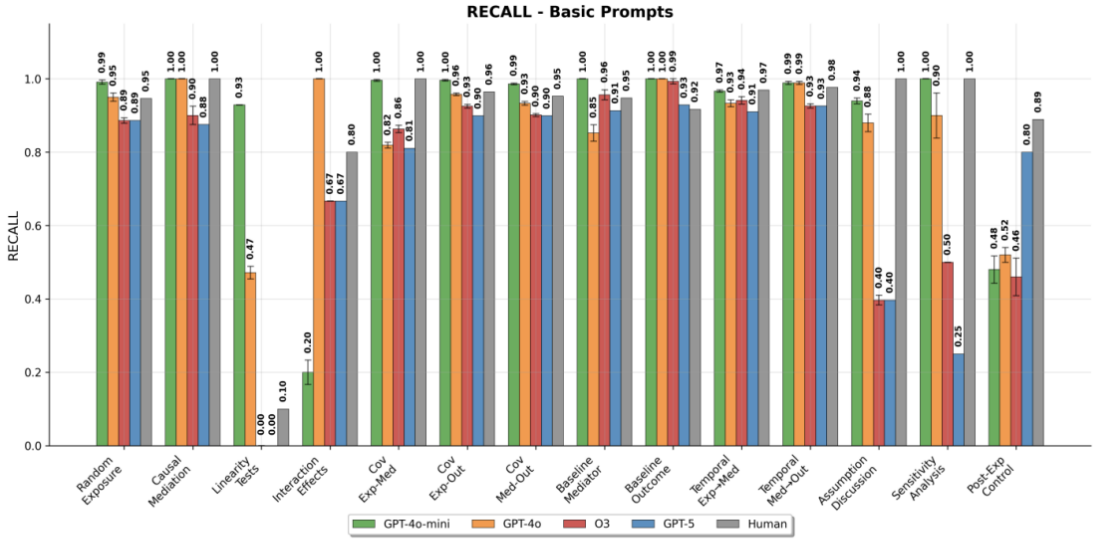
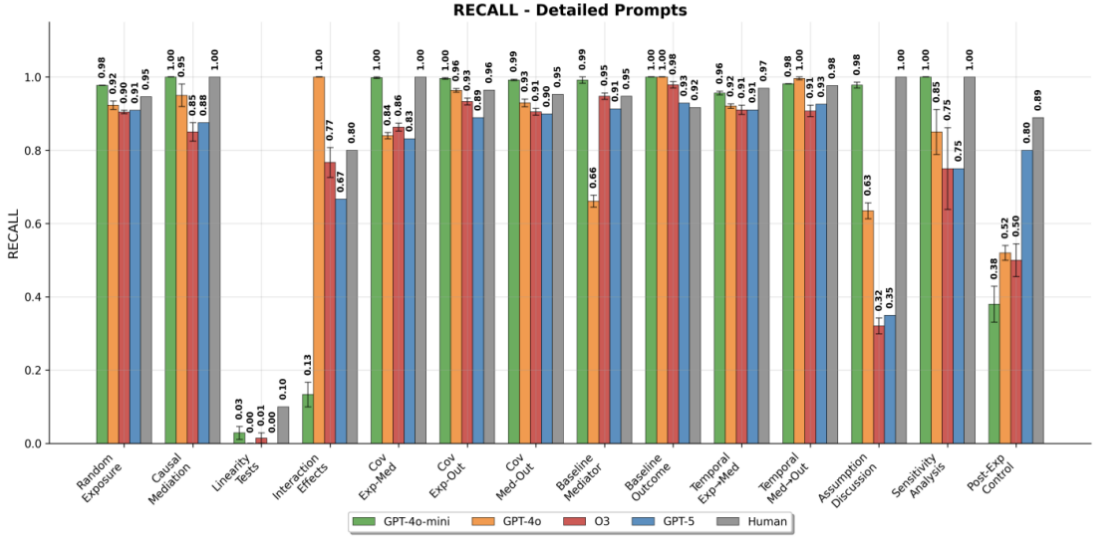

Figure S3. Model Recall Score Under Basic (Top) and Detailed (Bottom) Prompting Strategies Across 14 Methodological Criteria; colors represent different ChatGPT models and best human reviewer.

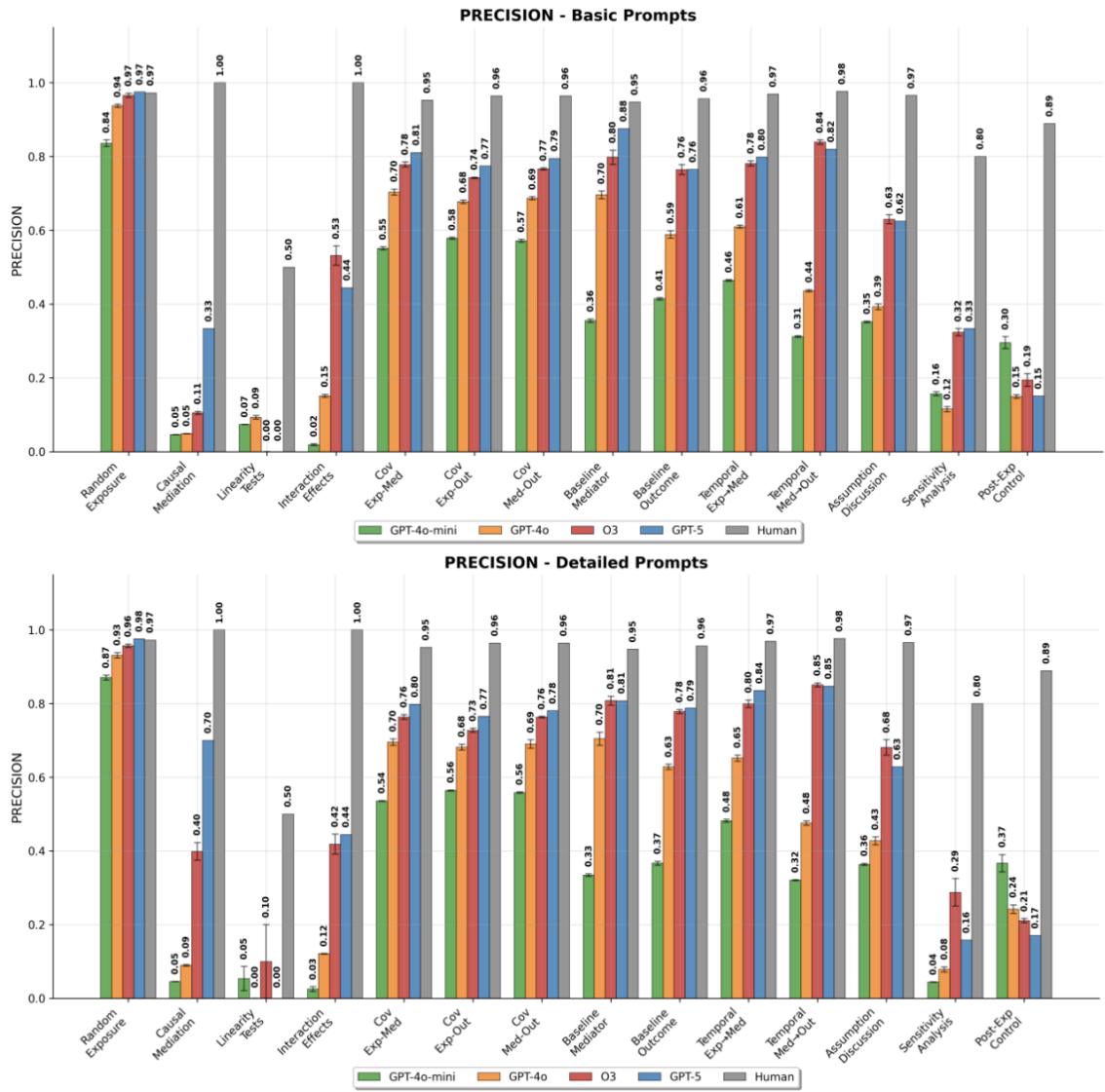

*Figure S4. Model Precision Score Under Basic (Top) and Detailed (Bottom) Prompting Strategies Across 14 Methodological Criteria; colors represent different ChatGPT models and best human reviewer.*

## Appendix A2: Additional Performance Metrics for Human Reviewers

There were seven human reviewers in the original Stuart et al.[7] study and there is substantial variation in their performance, especially when measured by F1 and precision (Figures S6 and Figure S7). Overall, similar to LLM results, less performant reviewers struggle with both low precision (more false positives) and low recalls (more false negatives).

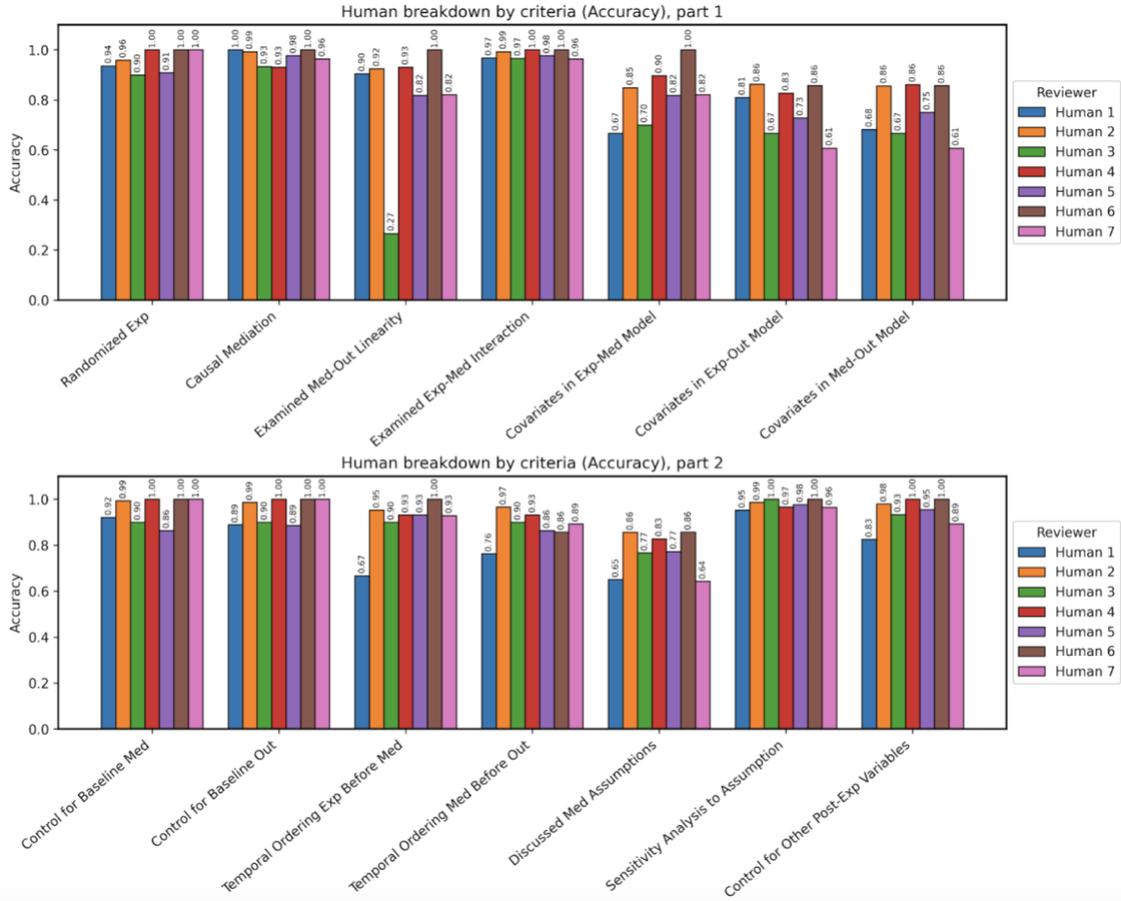

Figure S5. Human reviewer accuracy across 14 methodological criteria. Each reviewer was assigned a different number and set of papers.

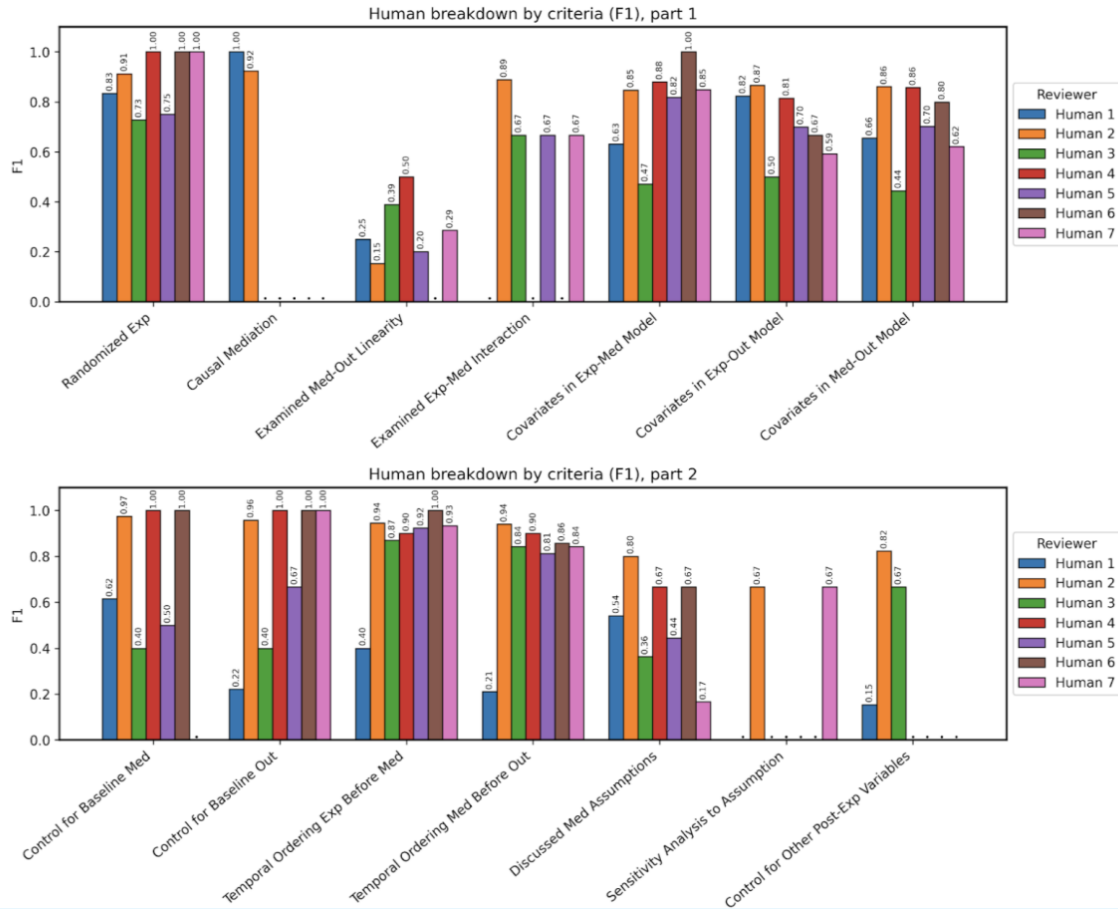

Figure S6. Human reviewer F1 score across 14 methodological criteria. Each reviewer was assigned a different number and set of papers.

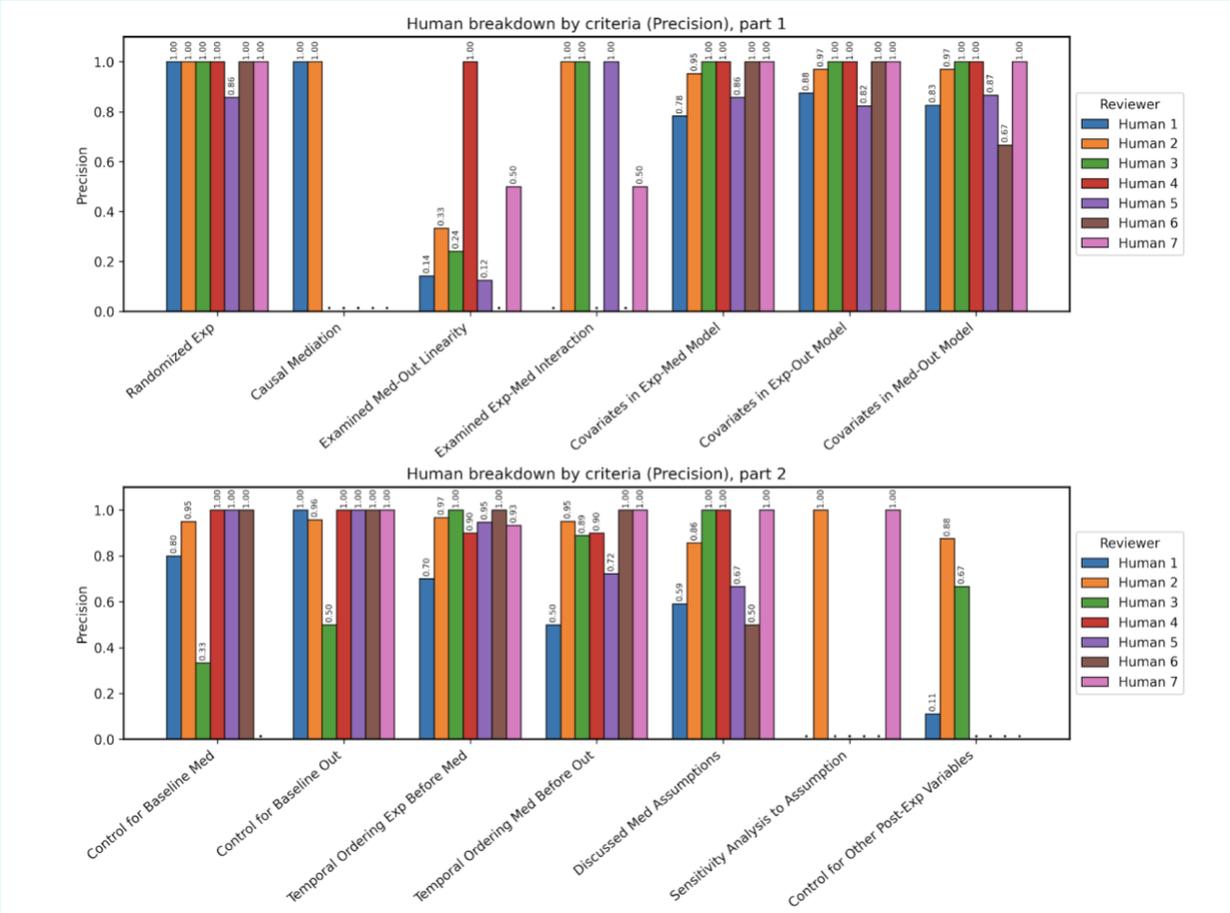

Figure S7. Human reviewer precision across 14 methodological criteria. Each reviewer was assigned a different number and set of papers.

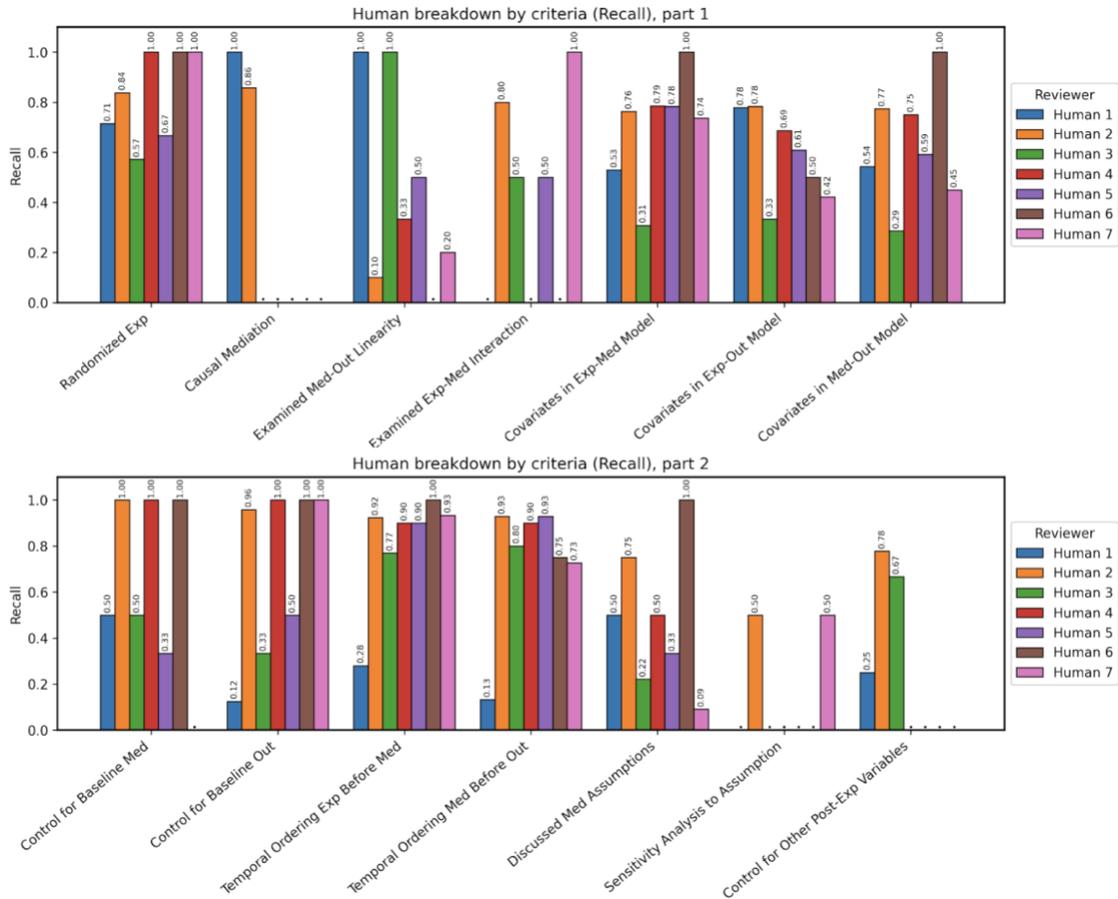

*Figure S8. Human reviewer recall across 14 methodological criteria. Each reviewer was assigned a different number and set of papers.*

**Appendix A3: Additional Models**

Here, we report performance results from Claude 4 and Gemini 2.5. Both models demonstrated performance patterns broadly similar to the GPT family, with accuracies closely matching those of GPT-5 and o3 across most criteria. These findings are consistent with extensive benchmarking efforts showing that these "state-of-the-art" models converge toward comparable capabilities in general language understanding and generation tasks.

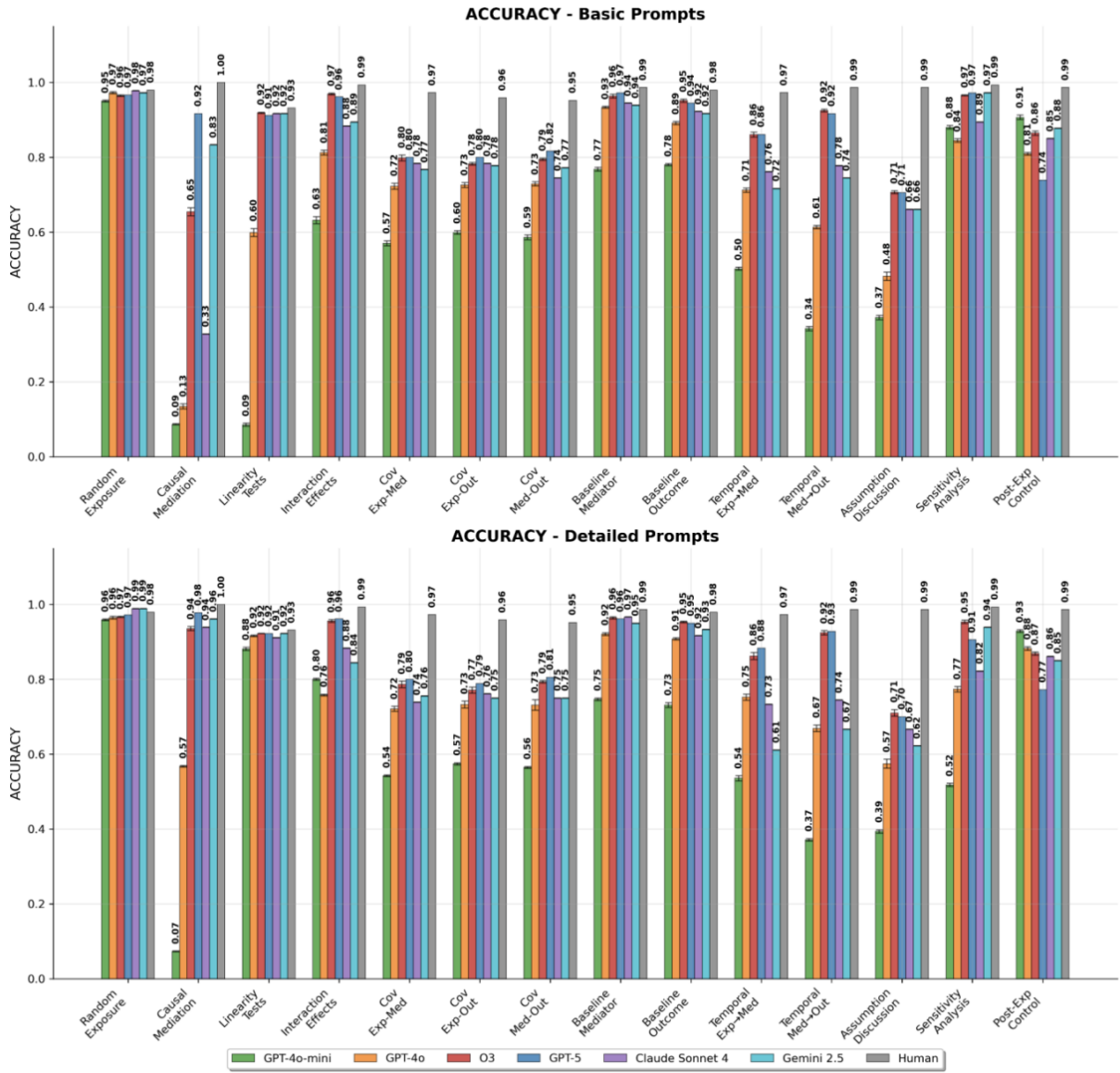

*Figure S9. Accuracy across 14 methodological criteria by prompting strategy (basic: top panel; detailed: bottom panel). Results include GPT-family models and two state-of-the-art non-GPT models for comparison.*

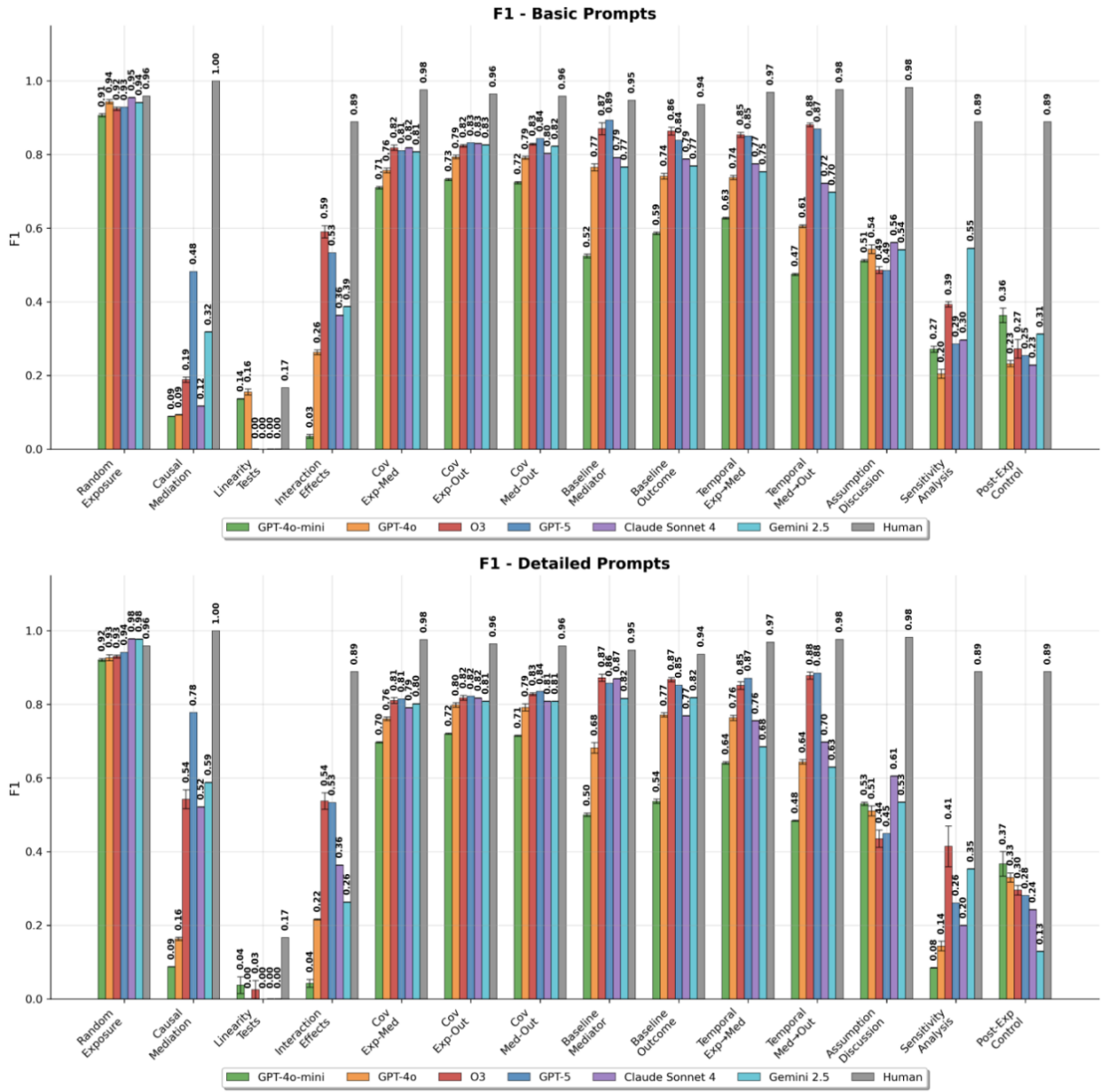

Figure S10. F1 across 14 methodological criteria by prompting strategy (basic: top panel; detailed: bottom panel). Results include GPT-family models and two state-of-the-art non-GPT models for comparison.

# Causal Mediation Extraction Prompt
Expert Instructions with Structured Output and Confidence Scores

> **BASIC**
>
> **Role.** You are a **causal inference expert** specialized in **mediation analysis**. Extract information from the article text.
>
> **Inputs**
> - **PMID:** {pmid}
> - **TEXT:** {text}
>
> **Fields to Extract**
> - **PMID:** [PMID]
> - **Title of the Study:** [title]
> - **Mediation Method Used:** [method name or "Not mentioned"]
>
> **Binary Fields (Answer with 1 or 0) & Confidence (0–1)**
> - Randomized Exposure: [0/1]   Confidence: [0–1]
> - Causal Mediation: [0/1]   Confidence: [0–1]
> - Examined Mediator–Outcome Linearity: [0/1]   Confidence: [0–1]
> - Examined Exposure–Mediator Interaction: [0/1]   Confidence: [0–1]
> - Covariates in Exposure–Mediator Model: [0/1]   Confidence: [0–1]
> - Covariates in Exposure–Outcome Model: [0/1]   Confidence: [0–1]
> - Covariates in Mediator–Outcome Model: [0/1]   Confidence: [0–1]
> - Control for Baseline Mediator: [0/1]   Confidence: [0–1]
> - Control for Baseline Outcome: [0/1]   Confidence: [0–1]
> - Temporal Ordering Exposure Before Mediator: [0/1]   Confidence: [0–1]
> - Temporal Ordering Mediator Before Outcome: [0/1]   Confidence: [0–1]
> - Discussed Mediator Assumptions: [0/1]   Confidence: [0–1]
> - Sensitivity Analysis to Assumption: [0/1]   Confidence: [0–1]
> - Control for Other Post–Exposure Variables: [0/1]   Confidence: [0–1]
>
> **Output Format (JSON-like markdown)**
>
> ```
> {
>   "pmid": "<PMID>",
>   "title": "<Title>",
>   "method": "<Method or Not mentioned>",
> ```

```
  "flags": {
    "randomized_exposure": {"value": 0/1, "confidence": 0.00-1.00},
    "causal_mediation": {"value": 0/1, "confidence": 0.00-1.00},
    "mediator_outcome_linearity": {"value": 0/1, "confidence": 0.00-1.00},
    "exposure_mediator_interaction": {"value": 0/1, "confidence": 0.00-1.00},
    "covars_em": {"value": 0/1, "confidence": 0.00-1.00},
    "covars_eo": {"value": 0/1, "confidence": 0.00-1.00},
    "covars_mo": {"value": 0/1, "confidence": 0.00-1.00},
    "baseline_mediator": {"value": 0/1, "confidence": 0.00-1.00},
    "baseline_outcome": {"value": 0/1, "confidence": 0.00-1.00},
    "temporal_e_before_m": {"value": 0/1, "confidence": 0.00-1.00},
    "temporal_m_before_o": {"value": 0/1, "confidence": 0.00-1.00},
    "assumptions_discussed": {"value": 0/1, "confidence": 0.00-1.00},
    "sensitivity_analysis": {"value": 0/1, "confidence": 0.00-1.00},
    "post_exposure_controls": {"value": 0/1, "confidence": 0.00-1.00}
  }
}
```

### Detailed Guidelines

1. **Randomized Exposure:** Look for terms like "randomized", "random assignment", "RCT", "experimental design", "intervention group vs control group".

2. **Causal Mediation:** Look for language like "causal effect", "causal mediation", "natural direct/indirect effects", "counterfactual", "potential outcomes".

3. **Linearity:** Check if they tested non-linearity (polynomial terms, splines) or explicitly assessed mediator–outcome functional form.

4. **Interaction:** Look for tests of whether the exposure–mediator relationship varies by another factor (moderation, interaction, effect modification).

5. **Covariates:** Verify if confounding variables were controlled in each model (EM, EO, MO).

6. **Baseline Control:** Adjustment for baseline/pre-treatment values of mediator/outcome.

7. **Temporal Ordering:** Confirm exposure measured before mediator, mediator before outcome (e.g., T1→T2→T3).

8. **Assumptions:** Discussion of mediation assumptions (no unmeasured confounding, sequential ignorability, consistency, positivity).

9. **Sensitivity Analysis:** Robustness checks, alternative specifications, bias/sensitivity analyses relevant to mediation assumptions.

10. **Post-Exposure Variables:** Any controls that occur after exposure but before mediator/outcome (note: controlling these can induce bias; mark as 1 if present).

## Positive (1 = Yes) Indicators

Below are example phrases that indicate a positive (1 = Yes) identification for each field.

- **Randomized Exposure:**
    - "Participants were randomly assigned to treatment or control"
    - "RCT design with randomization"
    - "Experimental manipulation of exposure"
    - "Random assignment to intervention groups"

- **Causal Mediation:**
    - "Causal mediation analysis"
    - "Natural direct and indirect effects"
    - "Counterfactual framework"
    - "Potential outcomes approach"
    - "Causal inference methods"

- **Examined Mediator–Outcome Linearity:**
    - "Tested for non-linear relationships"
    - "Polynomial terms included"
    - "Quadratic effects examined"
    - "Curvilinear relationship tested"

- **Examined Exposure–Mediator Interaction:**
    - "Moderation by third variable"
    - "Interaction effects tested"
    - "Conditional effects examined"
    - "Effect modification assessed"

- **Covariates in Models:**
    - "Adjusted for age, sex, education"
    - "Controlled for confounding variables"
    - "Covariates included in regression"
    - "Demographic variables controlled"

- **Baseline Control:**
    - "Baseline mediator controlled"
    - "Pre-treatment values adjusted"
    - "Initial levels accounted for"

- – "Baseline outcome included"
- **Temporal Ordering:**
    - – "Exposure measured at time 1"
    - – "Mediator measured at time 2"
    - – "Outcome measured at time 3"
    - – "Longitudinal design with proper timing"
- **Assumptions Discussion:**
    - – "Mediation assumptions discussed"
    - – "No unmeasured confounding assumption"
    - – "Sequential ignorability"
    - – "Limitations of mediation analysis"
- **Sensitivity Analysis:**
    - – "Sensitivity analysis conducted"
    - – "Robustness checks performed"
    - – "Alternative specifications tested"
    - – "Bias analysis conducted"
- **Post-Exposure Variables:**
    - – "Controlled for variables occurring after exposure"
    - – "Intermediate variables adjusted"
    - – "Post-treatment confounders included"

Template: Meta-style colored boxes using `tcolorbox`. Replace `{pmid}` and `{text}` with actual inputs.